\documentclass[conference]{IEEEtran}
\IEEEoverridecommandlockouts
\usepackage{cite}
\usepackage{amsmath,amssymb,amsfonts}
\usepackage{algorithmic}
\usepackage{graphicx}
\usepackage{textcomp}
\usepackage{xcolor}
\usepackage{booktabs}
\usepackage{multirow}
\usepackage{utfsym}
\usepackage[pagebackref=true,breaklinks=true,letterpaper=true,colorlinks,citecolor=blue,linkcolor=blue,bookmarks=false]{hyperref}

\def\BibTeX{{\rm B\kern-.05em{\sc i\kern-.025em b}\kern-.08em
    T\kern-.1667em\lower.7ex\hbox{E}\kern-.125emX}}
\begin{document}

\title{CS-PCN: Context-Space Progressive Collaborative Network for Image Denoising\\

\thanks{

\noindent*Corresponding author

\noindent Supported by Zhengtu Education of Beijing Jiaotong University Alumni Foundation, and Institute of Information Science, Beijing Jiaotong University, Beijing 100044, China, and also with the Beijing Key Laboratory of Advanced Information Science and Network Technology in the School of Computer, Beijing 100044, China.}
}

\author{\IEEEauthorblockN{ Yuqi Jiang}
\IEEEauthorblockA{\textit{Beijing Jiaotong University} \\
\textit{Beijing}\\
China \\
21125176@bjtu.edu.cn}
\and
\IEEEauthorblockN{ Chune Zhang*}
\IEEEauthorblockA{\textit{Beijing Jiaotong University} \\
\textit{Beijing}\\
China \\
chezhang@bjtu.edu.cn}
\and
\IEEEauthorblockN{ Jiao Liu}
\IEEEauthorblockA{\textit{Nankai University} \\
\textit{Tianjin}\\
China \\
jiaoliu@mail.nankai.edu.cn}
}

\maketitle

\begin{abstract}
Currently, image-denoising methods based on deep learning cannot adequately reconcile contextual semantic information and spatial details. To take these information optimizations into consideration, in this paper, we propose a Context-Space Progressive Collaborative Network (CS-PCN) for image denoising. CS-PCN is a multi-stage hierarchical architecture composed of a context mining siamese sub-network (CM2S) and a space synthesis sub-network (3S). CM2S aims at extracting rich multi-scale contextual information by sequentially connecting multi-layer feature processors (MLFP) for semantic information pre-processing, attention encoder-decoders (AED) for multi-scale information, and multi-conv attention controllers (MCAC) for supervised feature fusion. 3S parallels MLFP and a single-scale cascading block to learn image details, which not only maintains the contextual information but also emphasizes the complementary spatial ones. Experimental results show that CS-PCN achieves significant performance improvement in synthetic and real-world noise removal.
\end{abstract}


\section{Introduction}
Image denoising is a foundational task in computer vision, which aims to eliminate noises present from low-quality images and restore spatial details. Since many crucial features in the original image have been severely damaged by indeterministic noises \cite{martin2001database,dabov2007color,abdelhamed2018high,plotz2017benchmarking}, image denoising is still a challenging problem.

Image-denoising methods can be roughly divided into traditional methods and deep learning-based methods. Traditional methods \cite{dabov2007image,michaeli2013nonparametric,mairal2009non} primarily focus on image similarity with identifying patterns to remove noises. These techniques have made significant progress in removing Additive White Gaussian Noise (AWGN) \cite{dabov2007image,zhang2018ffdnet}. However, these methods do not effectively deal with the interference of real-world noise in the image. Compared with the traditional methods, deep learning-based methods \cite{zamir2020cycleisp,zhang2017learning, chen2022hider,fang2020multilevel} can efficiently learn abundant image priors by a sufficient number of image pairs. Especially, deep learning-based methods include single-stage networks and multi-stage networks. The performance gain of single-stage denoising methods is mainly attributed to the model design, such as encoder-decoders \cite{ronneberger2015u}, residual networks \cite{zhang2017beyond,tai2017memnet}, and generative adversarial networks (GAN) \cite{lv2021deep}. However, they tend to confront some limitations, including poor precision of extracted features
and insufficient network flexibility, etc. In contrast, multi-stage networks \cite{zhang2020msfsr,zamir2021multi, chen2021hinet} further address these problems and achieve superior results over single-stage ones. Nevertheless, existing hierarchy structures at each stage cannot exploit both global context and local detail information enough, which also limits the performance of multi-stage networks.

To address the above problems, in this paper, we propose a Context-Space Progressive Collaborative Network (CS-PCN) for image denoising. CS-PCN is a multi-stage hierarchical architecture and can be divided into two sub-networks to solve different sub-tasks. In particular, a context mining siamese sub-network (CM2S) is designed as a serial structure to capture multi-scale contextual information, which includes three parts: multi-layer feature processors (MLFP) for semantic information pre-processing, attention encoder-decoders (AED) for capturing multi-scale information, and multi-conv attention controllers (MCAC) for supervised feature fusion. To compensate for spatial details, we design a space synthesis sub-network (3S) with a parallel structure, consisting of the MLFP and a cascading block using dual attention blocks (DAB) and global average poolings (GAP). The 3S extracts the local spatial information that is carried out with the global context extraction. Extensive experiments show that CS-PCN produces promising results for image denoising. The main contributions of this work are:

\begin{itemize}
  \item We propose a Context-Space Progressive Collaborative Network (CS-PCN) with a multi-stage architecture. It coordinates the global contexts and local spaces for image denoising;
  \item We present two sub-networks at different stages, which are complementary to the other. CM2S relies on a serial siamese structure for multi-scale contextual learning, and 3S complements the spatial details with a parallel structure;
  \item We further introduce an attention control block at the end of each stage which can flexibly supervise feature aggregation and interaction.
\end{itemize}

\begin{figure*}[htbp]
\vspace{-6mm}
\centering
\includegraphics[width=0.95\textwidth]{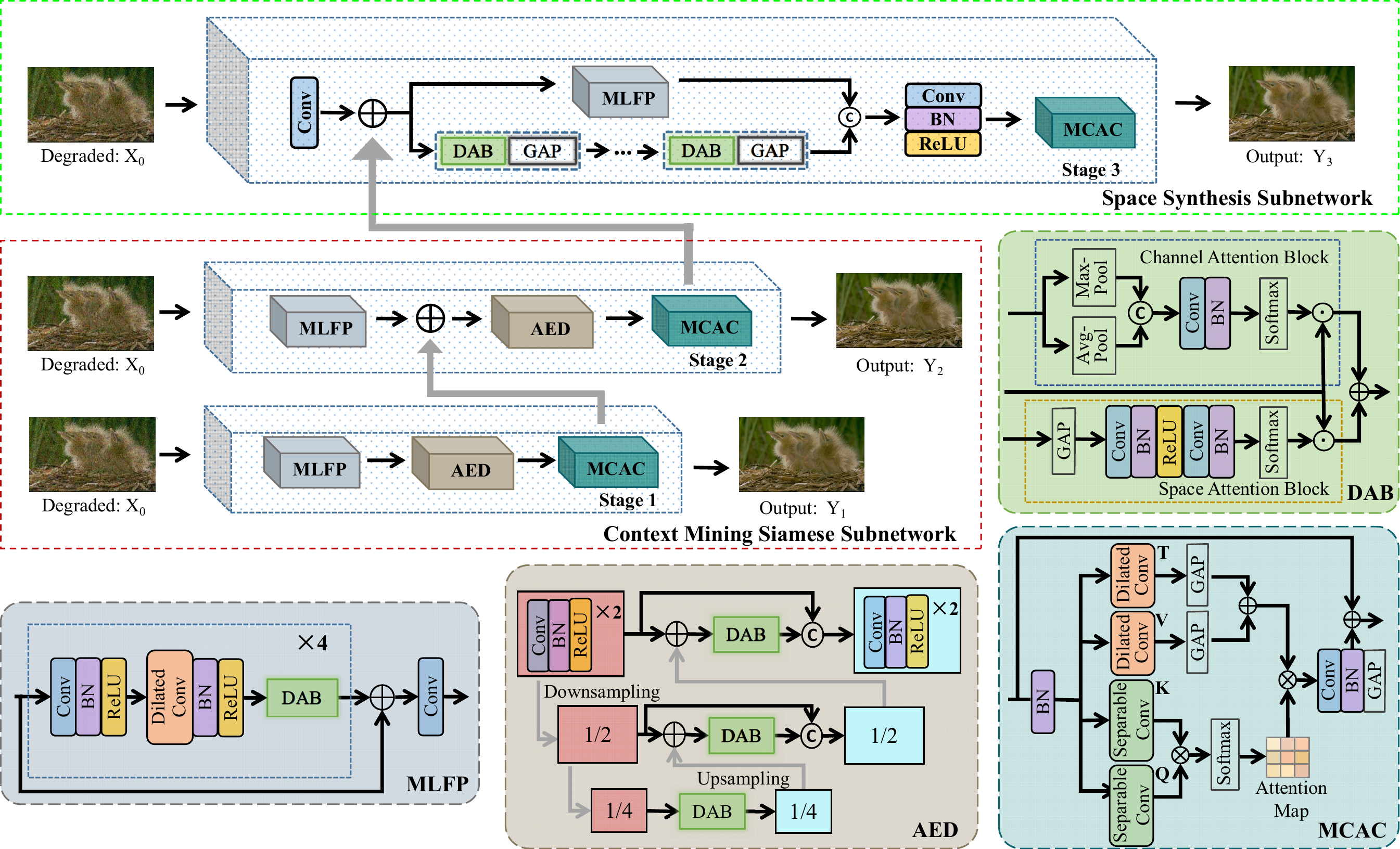}
\vspace{0mm}
\caption{ $\textbf{CS-PCN architecture.}$ The proposed network includes two sub-networks with three stages for progressive image denoising. Earlier stages employ CM2S to capture contextual information, and 3S is located at the last stage to generate spatial precise outputs.}
\vspace{-2mm}
\label{FIG:1}
\end{figure*}

\section{RELATED WORK}
\noindent \textbf{Single-Stage Network.}
Considering the inference speed, both traditional and deep learning-based denoising methods are mostly designed with a single-stage architecture. It only requires a single network to perform an image vision task. Particularly, DnCNN \cite{zhang2017beyond} was designed based on residual learning~\cite{liu2019dual, zhang2019residual,zamir2020learning} and batch normalization to achieve fast convergence and promising performance with a deep network. Chen \emph{et al.}~\cite{chen2018image} proposed to model the noise distribution using GAN \cite{lv2021deep} for blind denoising, which generated high-quality samples and guided subsequent denoising work. ADNet \cite{tian2020attention} proposed an attention-oriented \cite{anwar2019real} denoising network to integrate global and local features among long ranges, which improved the effectiveness of the model. However, single-stage networks only produce the final denoising image directly, it is not flexible enough and frequently leads to a lack of denoising precision when faced with complex noises.

\noindent\textbf{Multi-Stage Network.}
Multi-stage networks \cite{zhang2020msfsr,zamir2021multi, chen2021hinet} have gained more promising results than single-stage ones in image denoising, and have even been applied to other tasks, such as image deraining, deblurring \cite{kupyn2019deblurgan}, and super-resolution \cite{zhang2020msfsr}. Multi-stage networks decompose an entire network into manageable sub-networks to learn abundant features. The complex task is split into several simple sub-tasks to be completed sequentially. For example, MPRNet \cite{zamir2021multi} was presented with three serial stages to generate context-rich outputs, which was applied to different restoration tasks. MSPNet \cite{bai2023mspnet} employed a progressive pattern to remove the noise through parallel structural stages. However, adopting the same sub-structure at each stage would greatly limit the separate extraction for multi-faceted features and information aggregation. Based on the above shortcomings, our network model incorporates a specially designed diverse hierarchical architecture to improve denoising performance.

\section{Method} 
\subsection{Framework Overview}
Our CS-PCN consists of three stages to reconcile contextual and spatial information, which is shown in Fig.~\ref{FIG:1}. Given a degraded image $X_0 \in\mathbb{R}^{H\times W\times 3}$, where $H \times W$ denotes the height and the width of the image, CS-PCN applies CM2S to obtain a low-level feature $X_2 \in\mathbb{R}^{H\times W\times C}$ in the first two stages, where $C$ denotes the channel number. It can be formulated as: $ X_2 = CM2S\left( X_0 \right) $. Based on the serial structure, CM2S extracts multi-scale contextual information by enlarging the receptive field. CM2S includes several different modules, which are described in the following section. 

Notably, maintaining fine details is crucial for image denoising. Hence, we developed 3S with a parallel structure at the third stage, to exploit contextual and detailed information. This process is formulated as: $ Y_3 = 3S\left( X_0 \oplus X_2 \right) $, where $Y_3$ denotes the restored output, and $\oplus$ is element-wise addition. 3S is the context mining siamese sub-network that will be described in the next.

\subsection{Context Mining Siamese Sub-network}
Since convolutional neural networks (CNN) struggle to model dependencies over great distances, we configure the identical stage structure to form CM2S at the first two stages for enhancing the capture of global contextual information. CM2S adopts a serial structure to sequentially connect the designed MLFP, AED, and MCAC. The key designs are shown below.

\noindent \textbf{Multi-Layer Feature Processor (MLFP).}
We design an efficient feature processor with multi-convolutional layers, which performs a pre-processing on the features hidden in complex backgrounds, to provide excellent context information for further processing. We apply 3$\times$3 convolutions coupled with dilated ones to capture the global feature with a relatively wide range. Moreover, batch normalization is adopted behind each convolutional layer to reduce the complexity. We also utilize an attention mechanism (DAB) as a monitor to further improve the information flow.

\noindent \textbf{Attention Encoder-Decoder (AED).}
A novel encoder-decoder module based on U-Net \cite{ronneberger2015u} is designed to capture multi-scale contextual information by expanding the receptive field of MLFP. To suppress worthless information, DABs are adopted as the bottleneck of skip connections. Then the upsampling operation is applied to replace transposed convolution in the decoder to extend the feature resolution. Finally, we adopt the L1 loss function on decoder outputs to constrain the output features for its visual reality.

\noindent \textbf{Mult-Conv Attention Controller (MCAC).}
It has been proven in~\cite{han2020ghostnet} that implicit features contain information redundancy. Therefore, we introduce MCAC with a pyramid structure at the end of the stages. It has demonstrated superior capability in cross-stage transfer and feature fusion. We utilize separable convolutions consisting of depthwise and pointwise ones to perform pixel-by-pixel aggregation, and then learn inter-pixel dependencies by computing cross-covariance across features to get an attention map. This can capture spatial local information adequately and the feature map is obtained as:
\begin{equation}
Att (\hat{T},\hat{V},K,Q)= ( \hat{T} \oplus \hat{V}) \otimes Softmax\left(K \otimes Q \right),  
\end{equation}
where initially generated tensors are $T$, $V$, $K$ and $Q$, the $\hat{T}$, $\hat{V}$ are reshaping tensors, $\otimes$ is matrix multiplication.
Dilated convolutions with different dilation are used to consolidate global multi-scale information. The output of MCAC is formulated as:

\begin{equation}
 x_{out} = x_{in} \oplus P\left(f^1 Att\left( \hat{T} , \hat{V} , K , Q\right) \right),
\end{equation}
where $x_{in}$ is the input feature, $P (\cdot)$ represents the GAP, and $f^i (\cdot)$ refers to a combination symbol for convolution operations.

\subsection{Space Synthesis Sub-network}

At the third stage of the network, we employ 3S to learn the spatial local information. 3S is a parallel structure combined with MLFP and cascading block. The cascading block is designed with several DABs and GAPs, providing precise details and textures.

\noindent \textbf{Dual Attention Block (DAB).}
We study an attention topology in place of the convolutional block attention module (CBAM)~\cite{woo2018cbam}. The design assures that DAB focuses on both channel and space dimensions to learn generalized weights. And the channel attention block (CAB)~\cite{woo2018cbam, hu2018squeeze, liu2022spatial} and the spatial attention block (SAB)~\cite{woo2018cbam} are employed in parallel for processing. CAB generates a channel map ($M_C$) by calculating channel attention and information weights, formulated as:
\begin{equation}
 M_C= x_{in}\odot\sigma\left(f^2\delta\left(f^1 P\left(x_{in}\right)\right)\right) ,
\end{equation}
where $\delta (\cdot)$ and $\sigma (\cdot)$ represent the ReLU and Sigmoid activation function respectively, $\odot$ represents element-wise multiplication. SAB generates a spatial map ($M_S$), which can be formulated as:
\begin{equation}
 M_S\! = \!x_{in}\odot\sigma\left(f^1 C\left(Avgpool\left(x_{in} \right),Maxpool \left(x_{in} \right)\right)\right) ,  
\end{equation}
where $C (\cdot)$ represents the concatenation operation. Finally, $M_C$ and $M_S$ are added together to get the output of DAB. 

\subsection{Loss Function}
To mutually constrain the whole network, we adopt different loss functions on sub-networks: $S_1$, $S_2$, and $S_3$. Charbonnier loss, Edge loss, and reconstruction loss are adopted at the first two stages. Charbonnier loss and Edge loss are adopted at the third stage. Specifically, Charbonnier loss $L_{char}$ is formulated as:
\begin{eqnarray}
\begin{aligned}
L_{char} = \sqrt{\parallel X_{GT}-Y_{S=k} \parallel ^2 + \varepsilon ^2 } ,
\end{aligned}
\end{eqnarray}
where $X_{GT}$ denotes the ground truth, $Y_{S=k}$ (k=1,2,3) denotes the output image at the stage $k$ and $\varepsilon$ is experimentally set to $1 \times 10^{-3}$. Edge loss $L_{edge}$ is formulated as:
\begin{eqnarray}
\begin{aligned}
L_{edge} = \sqrt{\parallel\Delta\left(Y_{S=k}\right)-\Delta\left(X_{GT}\right) \parallel ^2 + \varepsilon ^2 } ,
\end{aligned}
\end{eqnarray}
where $\Delta$ represents the Laplace operator. Reconstruction loss $L_1$ is formulated as:
\begin{eqnarray}
\begin{aligned}
L_1 = \sum_{j = 1}^{2}\mid Y_j-X_{GT}\mid _1 ,
\end{aligned}
\end{eqnarray}
where $Y_j$ denotes the feature map of the $j$-th decoded output. The whole loss function is jointly formulated as: 
\begin{equation}
L\! =\!\sum_{S = 1}^{2}\left(L_{char}\!+\!\lambda _1 L_{edge}\!+\!\lambda _2 L_1\right)\!+\!\sum_{S= 3}^{3}\left(L_{char}\!+\!\lambda _1 L_{edge}\right) ,
\end{equation}
where $\lambda _1$ and $\lambda _2$ control the proportion of these loss functions.

\section{Experiments}
\subsection{Experimental Settings}

\begin{table*}[]
\begin{center}
\caption{Quantitative results with grayscale image denoising. Best and second best results are \textbf{highlighted} and \underline{underlined}. }
\label{Tab:1}
\resizebox{\textwidth}{23mm}{
\begin{tabular}{|c|c|c|c|c|c|c|c|c|c|c|c|}
\hline
Dataset    & $\sigma$ & BM3D \cite{dabov2007image} & RED \cite{mao2016image}  & DnCNN \cite{zhang2017beyond} & MemNet \cite{tai2017memnet}& IRCNN \cite{zhang2017learning} & FFDNet \cite{zhang2018ffdnet} & RIDNet \cite{anwar2019real} & RDN \cite{zhang2020residual}  & RDN+ \cite{zhang2020residual} & \textbf{Ours}  \\ 
\hline
\multirow{4}{*}{BSD68 \cite{martin2001database}} & 10 & 33.31 & 33.99 & 33.88 & -   & 33.74 & 33.76  & -      & 34.00 & $\underline{34.01}$ & $\boldsymbol{34.19}$ \\
                         & 30 & 27.76 & 28.50 & 28.36 & 28.43  & 28.26 & 28.39  & 28.54  & 28.56 & $\underline{28.58}$ & $\boldsymbol{28.67}$ \\
                         & 50 & 25.62 & 26.37 & 26.23 & 26.35  & 26.15 & 26.30  & 26.40  & 26.41 & $\underline{26.43}$ & $\boldsymbol{26.59}$ \\
                         & 70 & 24.44 & 25.10 & 24.90 & 25.09  & -     & 25.04  & 25.12  & 25.10 & $\underline{25.12}$ & $\boldsymbol{25.32}$ \\ \hline
\multirow{4}{*}{Kodak24} & 10 & 34.39 & 35.02 & 34.90 & -    & 34.76 & 34.81  & -    & 35.17 & $\underline{35.19}$ & $\boldsymbol{35.27}$ \\
                         & 30 & 29.13 & 29.77 & 29.62 & 29.72  & 29.53 & 29.70  & 29.90  & 30.00 & $\underline{30.02}$ & $\boldsymbol{30.14}$ \\
                         & 50 & 26.99 & 27.66 & 27.51 & 27.68  & 27.45 & 27.63  & 27.79  & 27.85 & $\underline{27.88}$ & $\boldsymbol{28.05}$ \\
                         & 70 & 25.73 & 26.39 & 26.08 & 26.42  & -     & 26.34  & 26.51  & 26.54 & $\underline{26.57}$ & $\boldsymbol{26.67}$ \\ \hline
\multirow{4}{*}{Urban100} & 10 & 34.47 & 34.91 & 34.73 & -    & 34.60 & 34.45  & -    & 35.41 & $\underline{35.45}$ & $\boldsymbol{35.52}$ \\
                         & 30 & 28.75 & 29.18 & 28.88 & 29.10  & 28.85 & 29.03  & -  & 30.01 & $\underline{30.08}$ & $\boldsymbol{30.10}$ \\
                         & 50 & 25.94 & 26.51 & 26.28 & 26.65  & 26.24 & 26.52  & -  & 27.40 & $\underline{27.47}$ & $\boldsymbol{27.66}$ \\
                         & 70 & 24.27 & 24.82 & 24.36 & 25.01  & -     & 24.86  & -  & 25.64 & $\underline{25.71}$ & $\boldsymbol{25.98}$ \\ \hline
\end{tabular}
}
\vspace{-2mm}
\end{center}
\end{table*}

\begin{table*}[]
\begin{center}
\caption{Quantitative results about color image denoising. Best and second best results are \textbf{highlighted} and \underline{underlined}.}
\label{Tab:2}
\resizebox{\textwidth}{23mm}{
\begin{tabular}{|c|c|c|c|c|c|c|c|c|c|c|c|}
\hline
Dataset   & $\sigma$ & CBM3D \cite{dabov2007color} & RED \cite{mao2016image}  & DnCNN \cite{zhang2017beyond} & MemNet \cite{tai2017memnet}& IRCNN \cite{zhang2017learning} & FFDNet \cite{zhang2018ffdnet} & RIDNet \cite{anwar2019real} & RDN \cite{zhang2020residual}  & RDN+ \cite{zhang2020residual} & \textbf{Ours}  \\ 
\hline
\multirow{4}{*}{BSD68 \cite{martin2001database}}  & 10 & 39.91 & 33.89 & 33.31 & -      & 36.06 & 36.14  & -      & 36.47 & $\underline{36.49}$ & $\boldsymbol{36.61}$ \\
                         & 30 & 29.73 & 28.46 & 30.40 & 28.39  & 30.22 & 30.31  & 30.47  & 30.67 & $\underline{30.69}$ & $\boldsymbol{30.78}$ \\
                         & 50 & 27.38 & 26.35 & 28.01 & 26.33  & 27.86 & 27.96  & 28.14  & 28.31 & $\underline{28.35}$ & $\boldsymbol{28.48}$ \\
                         & 70 & 26.00 & 25.09 & 26.56 & 25.08  & -     & 26.53  & 26.69  & 26.85 & $\underline{26.87}$ & $\boldsymbol{27.05}$ \\ \hline
\multirow{4}{*}{Kodak24} & 10 & 36.57 & 34.91 & 36.98 & -      & 36.70 & 36.81  & -      & 37.31 & $\underline{37.33}$ & $\boldsymbol{37.41}$ \\
                         & 30 & 30.89 & 29.71 & 31.39 & 29.67  & 31.24 & 31.39  & 31.64  & 31.94 & $\underline{31.98}$ & $\boldsymbol{32.06}$ \\
                         & 50 & 28.63 & 27.62 & 29.16 & 27.65  & 28.93 & 29.10  & 29.25  & 29.66 & $\underline{29.70}$ & $\boldsymbol{29.81}$ \\
                         & 70 & 27.27 & 26.36 & 27.64 & 26.40  & -     & 27.68  & 27.94  & 28.20 & $\underline{28.24}$ & $\boldsymbol{28.38}$ \\ \hline
\multirow{4}{*}{Urban100} & 10 & 36.00 & 34.59 & 36.21 & -    & 35.81 & 35.77  & -    & 36.69 & $\underline{36.75}$ & $\boldsymbol{36.79}$ \\
                         & 30 & 30.36 & 29.02 & 30.28 & 28.93  & 30.28 & 30.53  & -  & 31.69 & $\underline{31.78}$ & $\boldsymbol{31.80}$ \\
                         & 50 & 27.94 & 26.40 & 28.16 & 26.53  & 27.69 & 28.05  & -  & 29.29 & $\underline{29.38}$ & $\boldsymbol{29.43}$ \\
                         & 70 & 26.31 & 24.74 & 26.17 & 24.93  & -     & 26.39  & -  & 27.63 & $\underline{27.74}$ & $\boldsymbol{27.81}$ \\ \hline
\end{tabular}}
\vspace{-4mm}
\end{center}
\end{table*}

\noindent \textbf{Datasets and Metrics.}
We conduct experimental comparisons on synthetic and real-world noises. In synthetic noise experiments, we use the DIV2K dataset~\cite{agustsson2017ntire} with different levels of AWGN for training. The BSD68~\cite{martin2001database},  Kodak24, and Urban100 datasets are adopted for testing. In real-world noise experiments, the SIDD dataset \cite{abdelhamed2018high} is adopted for training and validation. The SIDD \cite{abdelhamed2018high} and DND~\cite{plotz2017benchmarking} datasets are adopted as test sets. The experimental performance is evaluated using PSNR and SSIM.


\noindent \textbf{Training Details.}
During training, random horizontal and vertical flipping are used for data augmentation. The patch size of inputs for synthetic and real-world noise experiments are set to $64\times 64$ and $128\times 128$ respectively. Batch size is set to $16$, the number of iteration is $4\times 10^5$, and ADAM optimizer \cite{kingma2014adam} is used to optimize the network. In synthetic denoising experiments, the initial learning rate is set to $1\times 10^{-4}$ and halved after every $1\times 10^5$ iterations. In real-world denoising experiments, the initial learning rate is set to $2\times 10^{-4}$ and the learning rate is steadily reduced to $1\times 10^{-6}$ by a cosine annealing strategy.

\subsection{Image Denoising Results}
\noindent \textbf{Synthetic Grayscale Denoising Experiments.}

\begin{figure}[!ht]
\centering
\includegraphics[width=0.48\textwidth, height=0.16\textwidth]{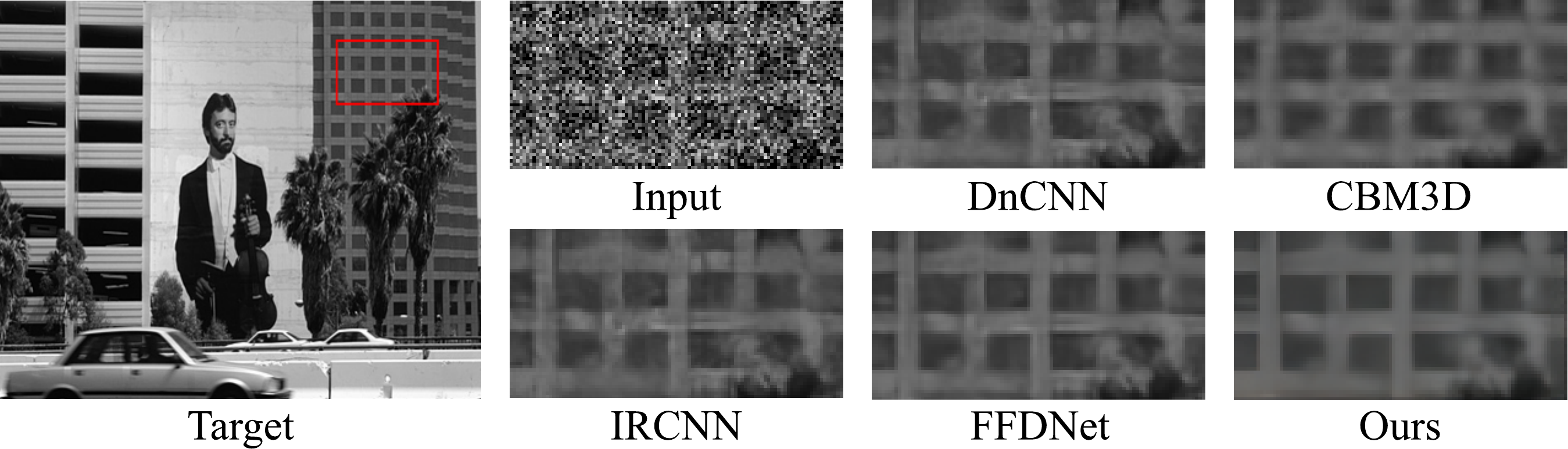}
\caption{Visualized denoised results of the synthetic grayscale images on the BSD68 dataset \cite{martin2001database} with noise level $\sigma$ = 50.}
\label{FIG:2}
\end{figure}

\begin{figure}[!h]
\centering
\includegraphics[width=0.48\textwidth, height=0.18\textwidth]{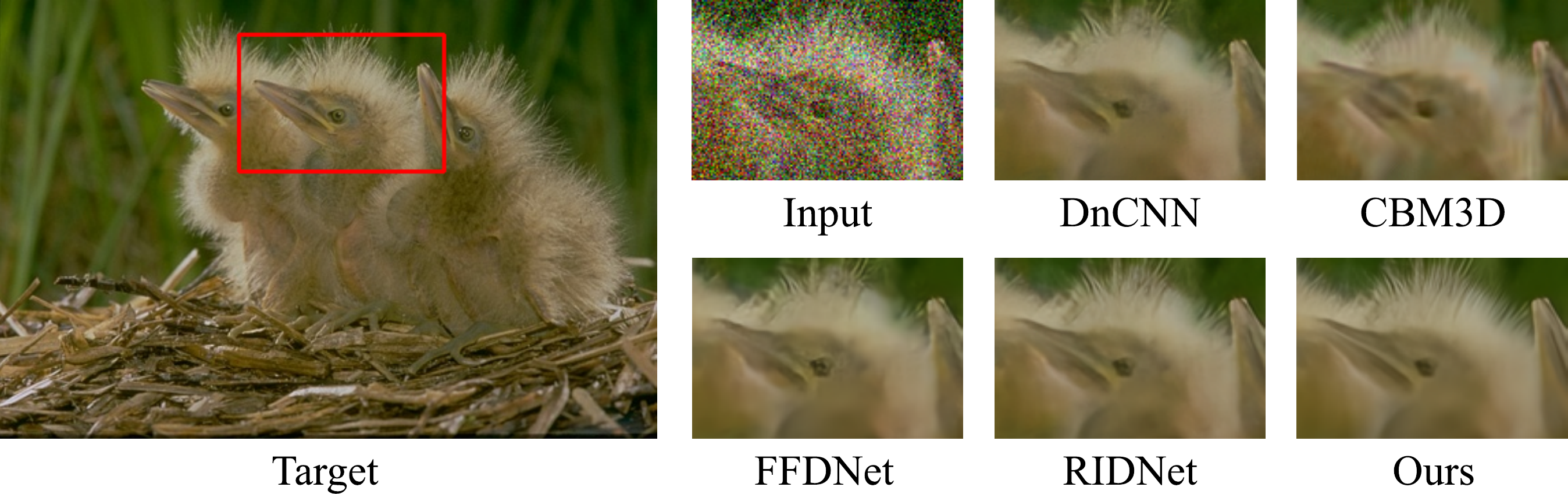}
\caption{Synthetic color image denoising results on the BSD68 dataset \cite{martin2001database} with noise level $\sigma$ = 50. The CS-PCN restores more perceptually-faithful images.}
\label{FIG:3}
\vspace{-4mm}
\end{figure}

Denoising results on BSD68 \cite{martin2001database}, Kodak24 and Urban100 datasets are summarized in Table~\ref{Tab:1}. It shows CS-PCN can achieve better results under grayscale noises of different levels. Compared with the SOTA algorithm RDN+ \cite{zhang2020residual}, we outperform it by 0.19 dB on the Urban100 dataset under the noise level $\sigma=50$. Qualitative comparisons are further shown in Fig.~\ref{FIG:2}. Although DnCNN \cite{zhang2017beyond}, and IRCNN \cite{zhang2017learning} remove noises to a certain extent, they also over-smooth some details. However, CS-PCN does better in capturing textures.


\begin{figure}[!h]
\centering
\includegraphics[width=0.48\textwidth, height=0.17\textwidth]{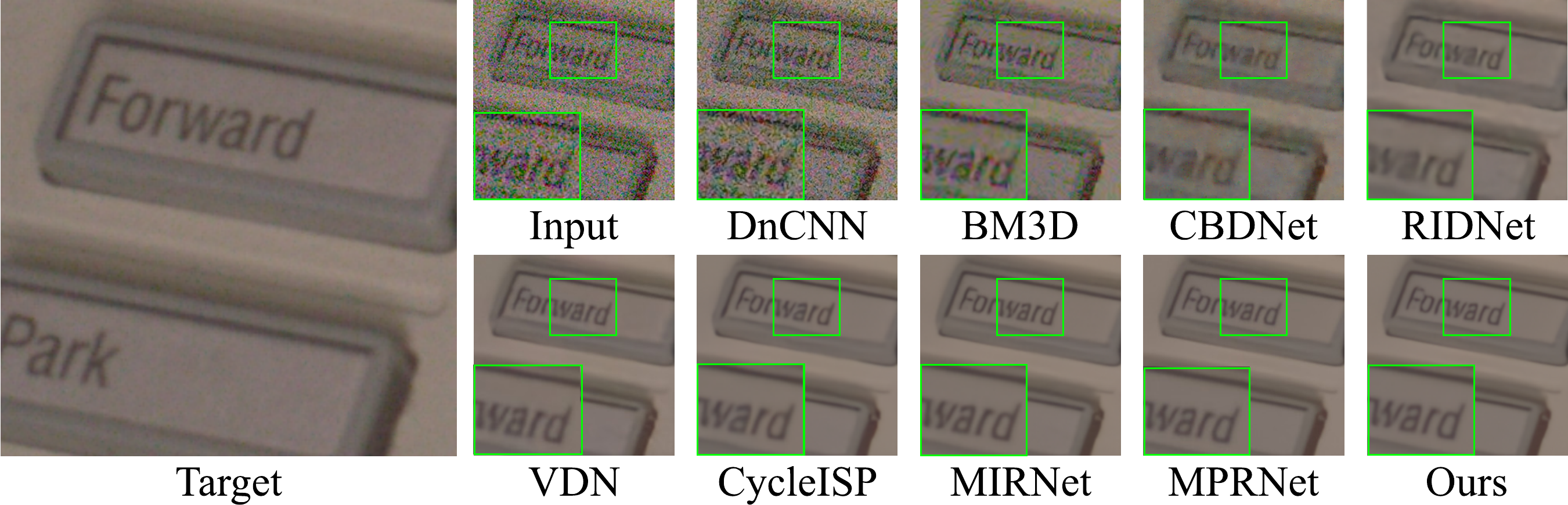}
\caption{The comparisons of real-world denoising from the SIDD dataset \cite{abdelhamed2018high} show that our model produces cleaner denoising results.}
\label{FIG:4}
\end{figure}

\begin{figure}[!ht]
\centering
\includegraphics[width=0.48\textwidth, height=0.17\textwidth]{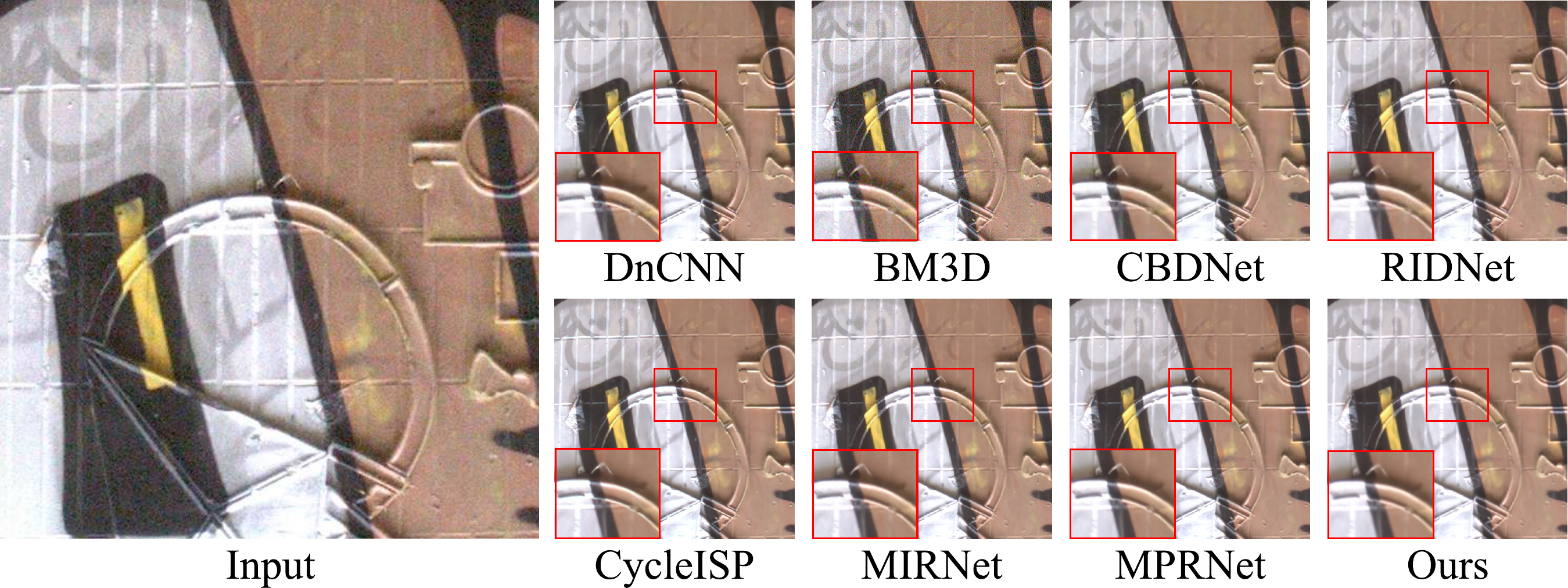}
\caption{The CS-PCN effectively removes real-world noises on the DND dataset \cite{plotz2017benchmarking} and is visually closer to the ground truth than others.}
\label{FIG:5}
\end{figure}

\begin{table}[!h]
\begin{center}
\caption{PSNR (dB) /SSIM results about real-world denoising on the SIDD \cite{abdelhamed2018high} and DND \cite{plotz2017benchmarking} datasets.}
\label{Tab:3}
\setlength{\tabcolsep}{1.0mm}{
\renewcommand\arraystretch{1.2}
\begin{tabular}{lll|ll|ll}
\hline\toprule 
         & \multicolumn{2}{c|}{SIDD \cite{abdelhamed2018high}} & \multicolumn{2}{c|}{DND \cite{plotz2017benchmarking}} & \multicolumn{2}{c}{Average} \\
Methods  & PSNR$\uparrow$        & SSIM$\uparrow$        & PSNR$\uparrow$        & SSIM$\uparrow$       & PSNR$\uparrow$         & SSIM$\uparrow$         \\ 
\midrule
DnCNN \cite{zhang2017beyond} & 23.66       & 0.583       & 32.43       & 0.790      & 28.04        & 0.686        \\
MLP \cite{burger2012image} & 24.71       & 0.641       & 34.23       & 0.833      & 29.47        & 0.737        \\
CBDNet \cite{guo2019toward} & 30.78       & 0.801       & 38.06       & 0.942      & 34.42        & 0.872        \\
BM3D \cite{dabov2007image} & 35.65       & 0.685       & 34.51       & 0.851      & 35.08        & 0.768        \\
RIDNet \cite{anwar2019real} & 38.71       & 0.951       & 39.26       & 0.953      & 38.99        & 0.952        \\
VDN \cite{yue2019variational}  & 39.28       & 0.956       & 39.38       & 0.952      & 39.33        & 0.954        \\
SADNet \cite{chang2020spatial} & 39.46       & 0.957       & 39.59       & 0.952      & 39.53        & 0.955        \\
CycleISP \cite{zamir2020cycleisp} & 39.52       & 0.957       & 39.56       & 0.956      & 39.54        & 0.957        \\
MPRNet \cite{zamir2021multi} & 39.71       & 0.958       & 39.80       & 0.954      & 39.76        & 0.956        \\
MIRNet \cite{zamir2020learning} & 39.72       & 0.959       & $\boldsymbol{39.88} $      & $\boldsymbol{0.959}$      & 39.80        & $\boldsymbol{0.959}$        \\
MAXIM-3S \cite{tu2022maxim} & $\underline{39.96}$  & $\underline{0.960}$ & 39.84       & 0.954      & $\underline{39.90}$ & 0.957        \\ 
\midrule
\textbf{Ours}   & $\boldsymbol{39.98}$     & $\boldsymbol{0.960}$ & $\underline{39.86}$ & $\underline{0.956}$ & $\boldsymbol{39.92}$  & $\underline{0.958}$        \\
\bottomrule\hline 
\end{tabular}}
\vspace{-4mm} 
\end{center}
\end{table}

\noindent \textbf{Synthetic Color Denoising Experiments.}
In Table~\ref{Tab:2}, the PSNR scores of several experiments are reported. Our model obtains considerable gains, 1.1 dB over CBM3D \cite{dabov2007color} on the BSD68 dataset \cite{martin2001database} of $\sigma=50$ and 0.71 dB over FFDNet \cite{zhang2018ffdnet} on the Kodak24 dataset of $\sigma=50$. And Fig.~\ref{FIG:3} qualitatively presents the intuitive denoising performance of different networks. It shows that our CS-PCN restores better edge details than others.

\noindent \textbf{Real-World Denoising Experiments.}
As observed in Table~\ref{Tab:3}, CS-PCN performs better than most CNN-based denoising networks. On the SIDD dataset \cite{abdelhamed2018high}, CS-PCN outperforms 0.27 dB over MPRNet \cite{zamir2021multi} which also uses the multi-stage architecture. Fig.~\ref{FIG:4} and Fig.~\ref{FIG:5} illustrate visual results on both datasets. 
\begin{table}[!h] 
\begin{center}
\caption{Evaluations of the number of stages on the SIDD dataset \cite{abdelhamed2018high}.}
\label{Tab:4}
\vspace{0mm}
\renewcommand\arraystretch{1.3}
\begin{tabular}{|c|c|c|c|}
  \hline
  Model Stage  & 1 & 2 & 3
  \\
  \hline
  PSNR & 39.65  & 39.79 & \textbf{39.98} \\
  \hline
\end{tabular}
\vspace{-4mm}
\end{center}
\end{table}
\begin{table}[!ht]  
\begin{center}
\caption{Ablation study on different components of the proposed CS-PCN on the BSD68 dataset \cite{martin2001database}.}
\label{Tab:5}
\vspace{0mm}
\setlength{\tabcolsep}{2.4mm}{
\renewcommand\arraystretch{1.3}
\begin{tabular}{lccc}
\hline\toprule 
Combination Schemes   &Stages      & MCAC & PSNR           \\ \hline
AED           &1              & -    & 28.16          \\
MLFP          &1              & -    & 28.18          \\
3S'(AED + GAP)    &1   & -   &  28.19       \\
3S''(AED + Cascading)   &1   & -  &28.21          \\
3S (MLFP + Cascading)          &1              & -    & 28.23     \\\hline
CM2S          &2              & $\times$    & 28.27          \\ 
3S + 3S          &2             & $\times$    & 28.30          \\
(MLFP + AED) + 3S     &2          & $\times$    & 28.32          \\ \hline
(MLFP + AED) + 3S + 3S      &3      & $\times$     & 28.35          \\
CM2S + 3S         &3            & $\times$     & 28.38          \\
CM2S + 3S       &3              & $\checkmark$     & \textbf{28.42} \\ \bottomrule\hline 
\end{tabular}}
\vspace{-4mm}
\end{center}
\end{table}
It can be observed that previous models cannot restore the texture areas well, and the image background also appears partially fused. CS-PCN obtains a great improvement in the subjective quality of images, which demonstrates that our model gets a preliminary solution to the aforementioned problems.

\subsection{Ablation Studies}

\noindent \textbf{Number of stages.}
As shown in Table~\ref{Tab:4}, our model performance is getting better as increasing the number of stages. CS-PCN can achieve an initial denoising effect in the first stage, while the second and third stages gradually improve network performance based on the first stage.

\noindent \textbf{Module combinations.}
In Table~\ref{Tab:5}, we demonstrate the combinatorial effects of different modules. It confirms the importance of each individual component and the best way to structure sub-networks (an improvement of around 0.1dB over three stages using the same sub-structures). To verify the validity of MCAC at the end of each stage, we remove it and the performance of our CS-PCN decreased. These results provide further evidence that our network combination scheme is scientific.

\section{CONCLUSIONS}
In this paper, a Context-Space Progressive Collaborative Network (CS-PCN) is designed for image denoising. It is a three-stage denoising model with two sub-networks. 

These two sub-networks extract contextual and spatial information in a series and parallel way respectively. And our network fully reconciles the features between contextual semantics and spatial details by the complementary sub-networks. Experimental results on many datasets demonstrate the superiority of CS-PCN over existing methods, in terms of objective and subjective performance. In the future, we will try to employ the network on more computer vision tasks and continue to explore lighter and more efficient models.

\bibliographystyle{IEEEtran}
\bibliography{IEEEexample}

\end{document}